\documentclass[mathpazo]{csiam-am}
\renewcommand\thisnumber{x}
\renewcommand\thisyear {20xx}
\renewcommand\thismonth{xx}
\renewcommand\thisvolume{x}

\usepackage[most]{tcolorbox}
\usepackage{xcolor}
\usepackage{enumitem}
\usepackage[utf8]{inputenc} 
\usepackage[T1]{fontenc}    
\usepackage{hyperref}       
\usepackage{url}           
\usepackage{booktabs}       
\usepackage{amsfonts}       
\usepackage{nicefrac}       
\usepackage{microtype}      
\usepackage{xcolor}         
\usepackage{graphicx}
\usepackage{listings}
\usepackage{booktabs} 
\usepackage{subcaption}
\usepackage{caption}
\usepackage{color}
\usepackage{amsmath}
\usepackage{cleveref}
\definecolor{keywordcolor}{rgb}{0.7, 0.1, 0.1}   
\definecolor{tacticcolor}{rgb}{0.0, 0.1, 0.6}    
\definecolor{commentcolor}{rgb}{0.4, 0.4, 0.4}   
\definecolor{symbolcolor}{rgb}{0.0, 0.1, 0.6}    
\definecolor{sortcolor}{rgb}{0.1, 0.5, 0.1}      
\definecolor{attributecolor}{rgb}{0.7, 0.1, 0.1} 

\title{REAL-Prover: Retrieval Augmented Lean Prover for Mathematical Reasoning}

\author[Z.~Shen, N.~Huang, F.~Yang, Y.~Wang, G.~Gao et~al.]{Ziju Shen\affil{1}\comma\footnotemark[2], Naohao Huang\affil{2}\comma\footnotemark[2], Fanyi Yang\affil{3}\comma\footnotemark[2], Yutong Wang\affil{1}\comma\footnotemark[2], 
Guoxiong Gao\affil{1}\comma\footnotemark[2], Tianyi Xu\affil{1}, Jiedong Jiang\affil{1}, Wanyi He\affil{1}, Pu Yang\affil{1}, Mengzhou Sun\affil{1}, Haocheng Ju\affil{1}, Peihao Wu\affil{3}, Bryan Dai\affil{3} and Bin Dong\affil{4}\comma\affil{5}\comma\affil{6}\comma\corrauth}
\address{\affilnum{1}\ Peking University\\
\affilnum{2}\ Renmin University of China\\
\affilnum{3}\ Ubiquant\\
\affilnum{4}\ Beijing International Center for Mathematical Research and the New Cornerstone Science Laboratory, Peking University\\
\affilnum{5}\ Center for Machine Learning Research, Peking University\\
\affilnum{6}\ Center for Intelligent Computing, Great Bay Institute for Advanced Study, Great Bay University\\
\affilnum{\footnotemark[2]}\ Equal contribution.\\
Code: \href{https://github.com/frenzymath/REAL-Prover}{https://github.com/frenzymath/REAL-Prover}
}
\emails{{\tt dongbin@math.pku.edu.cn}}
\begin{document}


\begin{abstract}
     Nowadays, formal theorem provers have made monumental progress on high-school and competition level mathematics, but few of them generalize to more advanced mathematics. In this paper, we present REAL-Prover, a new open-source stepwise theorem-prover for Lean 4 to push this boundary. This prover, based on our fine-tuned large language model (REAL-Prover-v1) and integrated with a retrieval system (Leansearch-PS), notably boosts performance on solving college-level mathematics problems. To train REAL-Prover-v1,  we developed HERALD-AF, a data extraction pipeline that converts natural language math problems into formal statements, and a new open-source Lean 4 interactive environment (Jixia-interactive) to facilitate synthesis data collection. In our experiments, our prover using only supervised fine-tuning achieves competitive results with a 23.7\% success rate (Pass@64) on the ProofNet dataset—comparable to state-of-the-art (SOTA) models. To further evaluate our approach, we introduce FATE-M, a new benchmark focused on algebraic problems, where our prover achieves a SOTA success rate of 56.7\% (Pass@64).
\end{abstract}
\maketitle

\section{Introduction}
The formalization of mathematics has long been a great challenge at the intersection of programming and mathematical logic. Over the past decades, theorem provers have made steady progress. In particular, proof assistants have been used to formalize complex theorems such as the Four-Color Theorem \cite{fourcolourthm} and the Feit-Thompson Odd-Order Theorem \cite{oddthm}. These achievements underscore the potential of formal methods to resolve research-level proofs.

Modern Interactive Theorem Provers (ITPs) like Lean \cite{lean} and Coq \cite{coq} have significantly improved usability and automation, lowering the barrier for adoption. Lean, for example, provides a user-friendly environment and a large community-driven library of mathematics (Mathlib) \cite{The_mathlib_Community_2020}. In early 2025, Lean's Mathlib had grown to encompass more than 210,000 formally proven theorems in various domains. Meanwhile, utilities like LeanSearch \cite{gao2025semanticsearchenginemathlib4} also emerge to help users navigate easily in these numerous domains and facts, saving their efforts on formal proof drafting.


\begin{figure}[t]
    \centering
        \includegraphics[width=\textwidth]{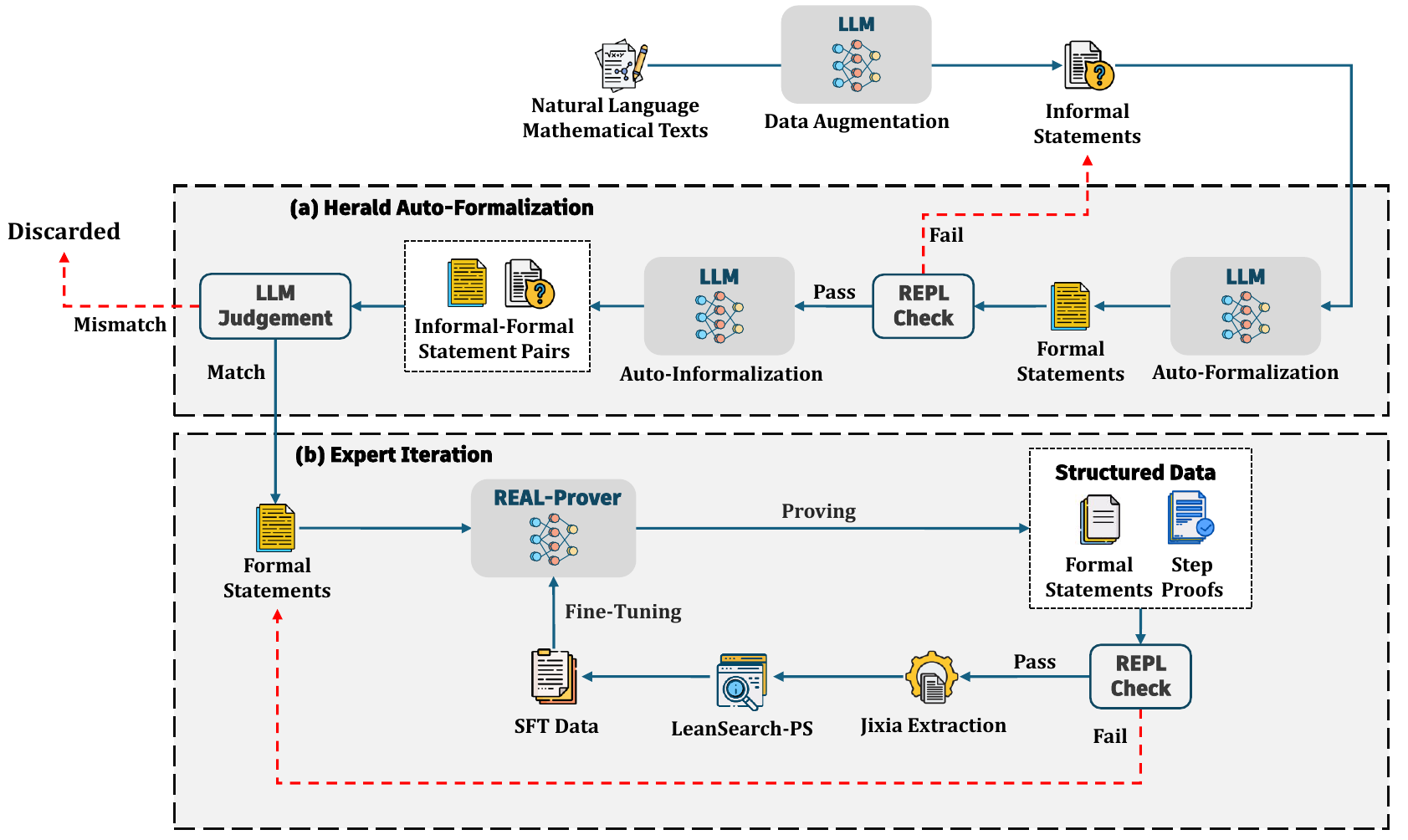}
        \caption{\textbf{Overview of the REAL-Prover model training pipeline.} The figure illustrates two components: (a) the HERALD-AF pipeline, which translates informal mathematical statements into formal ones; (b) the expert iteration pipeline, which iteratively refines the prover model. }
        \label{fig:overview_fig}
\end{figure}

Recent advances in large language models (LLMs) have significantly advanced automated theorem proving, successfully generating formal proofs for high-school-level math competitions \cite{polu2020generativelanguagemodelingautomated, jiang2023draftsketchproveguiding, zhao2023decomposingenigmasubgoalbaseddemonstration, wang2023legoproverneuraltheoremproving, xin2024deepseekprover, wu2024internlm2, xin2024deepseekproverv15harnessingproofassistant,xin2025bfsproverscalablebestfirsttree, ren2025deepseekproverv2advancingformalmathematical}. Nonetheless, extending these approaches to undergraduate and graduate-level mathematics remains a significant challenge due to data sparsity, the intricacies of domain-specific reasoning, and the requirement for precise retrieval and application of extensive lemma libraries, where even minor inaccuracies can compromise the correctness of purely neural proof generation.

To address these challenges, we propose \textbf{REAL-Prover} (REtrieval-Augmented Lean Prover), a retrieval-augmented and stepwise framework for automated theorem proving in Lean. Each model invocation is conditioned on the current proof state and enhanced through the retrieval of semantically relevant theorems from Mathlib. The framework comprises three key components. \textbf{LeanSearch-PS} performs semantic premise selection through a two-stage training strategy that improves both retrieval precision and generalization. \textbf{HERALD-AF} extends the HERALD pipeline to systematically translate informal mathematical statements into verified Lean expressions, enabling scalable data generation. \textbf{Jixia-Interactive} provides a Lean 4 interactive environment that support efficient interaction between prover and compiler in stepwise proving. For evaluation, we construct \textbf{FATE-M} (Formal Algebra Theorem Evaluation–Medium), a benchmark designed for graduate-level reasoning in algebraic domains including groups, rings, and fields. REAL-Prover achieves a state-of-the-art accuracy of 56.7\% on FATE-M and demonstrates competitive performance on miniF2F and ProofNet, confirming its effectiveness in complex formal reasoning.

\subsection{Related works}
\paragraph{Automated Theorem Proving}Automated Theorem Proving in formal languages, particularly within the framework of ITPs, is a rapidly advancing research area in artificial intelligence. LLM-based provers can be broadly categorized by their proof generation strategies: whole-proof generation that aims to produce complete proof in a single pass, and proof-step generation that iteratively refines proofs through ITPs.

The whole-proof generation paradigm focuses on end-to-end synthesis of formal proof scripts from theorem statements. Early attempts like DeepSeek-Prover-V1 \cite{xin2024deepseek} demonstrated the feasibility of this approach in Lean 4. Subsequent improvements in DeepSeek-Prover-V1.5 \cite{xin2024deepseekproverv15harnessingproofassistant} introduced hybrid techniques combining Reinforcement Learning from Proof Assistant Feedback (RLPAF) with Group Relative Policy Optimization \cite{shao2024deepseekmath}, along with a `truncate-and-resume' mechanism that allows error correction through partial stepwise verification. Goedel-Prover \cite{lin2025goedel} addressed data scarcity through auto-formalization, first converting natural language mathematics into formal statements before generating proofs through iterative training. While these approaches benefit from large-scale supervised fine-tuning and reduced ITP interaction, they struggle with error accumulation in long logical chains and implicit state inference.

This limitation has driven the development of interactive proof-step generation approaches that leverage ITPs feedback during the proving process. Systems like BFS-Prover \cite{xin2025bfsproverscalablebestfirsttree} employ expert iteration with Best-First Search, using Lean compiler feedback to guide tactic generation. HunyuanProver \cite{li2024hunyuanprover} enhances search efficiency through specialized critic models that evaluate intermediate proof states. The LeanDojo framework introduces ReProver \cite{yang2023leandojotheoremprovingretrievalaugmented}, which adopts a distinct retrieval-based approach using Dense Passage Retriever \cite{karpukhin2020dense} to directly predict relevant premises. LeanAgent \cite{kumarappan2024leanagent} extends this paradigm through lifelong learning, maintaining a dynamic knowledge base to progressively master theorem proving across multiple repositories. These interactive methods typically combine expert iteration, enabling real-time adaptation to proof state changes while managing computational complexity through guided tree search.

\paragraph{Formal Statement Benchmark}A fundamental aspect of developing and evaluating theorem provers is the utilization of formal statement benchmarks. Prominent examples include miniF2F \cite{miniF2F}, which offers a curated set of Olympiad-level formalization problems, and ProofNet \cite{azerbayev2023proofnetautoformalizingformallyproving}, which targets undergraduate-level mathematical reasoning. These benchmarks serve as standardized tools to assess the capabilities of theorem provers in generating correct and complete proofs. As the proficiency of provers advances, there is an increasing need for more challenging benchmarks, such as graduate-level or research-oriented problems, to rigorously evaluate their reasoning depth, adaptability, and ability to handle complex or previously unseen formalizations.

\subsection{Our Contributions}
Our main contributions are summarized as follows: 
\begin{itemize}[leftmargin=*]
\item \textbf{Retrieval-Augmented Proof System.} We present a unified proof system that integrates data collection, training, and inference. The system includes three main components. HERALD-AF provides a pipeline for translating informal mathematical statements into formal Lean expressions. REAL-Prover implements a stepwise theorem-proving framework with LeanSearch-PS for semantic premise selection and employs an expert-iteration training paradigm. Jixia-Interactive  facilitates efficient and stable interactions with the Lean 4 prover during both training and inference. The integration of these components significantly enhances proof performance on graduate-level formal mathematics.

\item \textbf{State-Tactic Pair Dataset.} We release a comprehensive dataset consisting of 55k state–tactic pairs generated through our expert iteration process. Each data instance encapsulates a formal proof state, its surrounding context, and the corresponding tactic sequence, providing high-quality supervision for both proof search optimization and policy learning.

\item \textbf{Specialized Benchmark for Abstract Algebra.} We present FATE-M (Formal Algebra Theorem Evaluation–Medium), a specialized benchmark comprising paired informal-formal problems in graduate-level abstract algebra. FATE-M complements existing benchmarks such as miniF2F \cite{miniF2F} and ProofNet \cite{azerbayev2023proofnetautoformalizingformallyproving}, establishing an effective framework for rigorous formal reasoning assessment.

\end{itemize}

\section{Method}
To construct a high-performance automated prover, we propose REtrieval-Augmented Lean Prover (REAL-Prover), a stepwise proof system for Lean 4. To supply REAL-Prover with a high-quality training corpus, we first introduce Hierarchy and Retrieval-based Translated Lean Dataset Auto-formalization (HERALD-AF) in \Cref{subsec:herald-af}, a pipeline that translates informal mathematical problems into formal Lean 4 statements. HERALD-AF converts informal mathematical problems into formal Lean 4 statements, yielding a large corpus of candidate theorems. Since these statements still lack certified proofs, in \Cref{subsec:real-prover}, we build a complementary pipeline—again driven by REAL-Prover—that synthesizes and automatically verifies the required proofs, completing the training loop.

\subsection{HERALD-AF}
\label{subsec:herald-af}

Our formal statement generation framework builds upon the Hierarchy and Retrieval-based Translated Lean Dataset (HERALD) \cite{gao2025heraldnaturallanguageannotated}, which establishes a protocol translating informal statements into formal statements. \Cref{fig:overview_fig}(a) schematizes HERALD-AF pipeline.

Before applying HERALD-AF, we perform a preprocessing step to extract informal statements from a natural language mathematics corpus. This begins with the use of regular expressions to identify theorem blocks and exercises within the text. Subsequently, an LLM is employed to generate and refine the extracted informal statements.

Once the informal statements are obtained, we apply HERALD-AF to translate them into formal statements. This translation pipeline comprises three key stages: Auto-Formalization, Auto-Informalization, and LLM Judgement. In the Auto-Formalization stage, an open-source model Herald-translator \cite{gao2025heraldnaturallanguageannotated} is used to generate multiple candidate formal statements from each informal input. During Auto-Informalization, these formal candidates are translated back into natural language using DeepSeek-V3 \cite{deepseekai2025deepseekv3technicalreport}. Finally, in the LLM Judgement stage, a general LLM evaluates whether each back-translated statement faithfully reflects the original input. Only those formal statements that pass this consistency check are retained.

\begin{figure}[t]
        \centering
        \includegraphics[width=\textwidth]{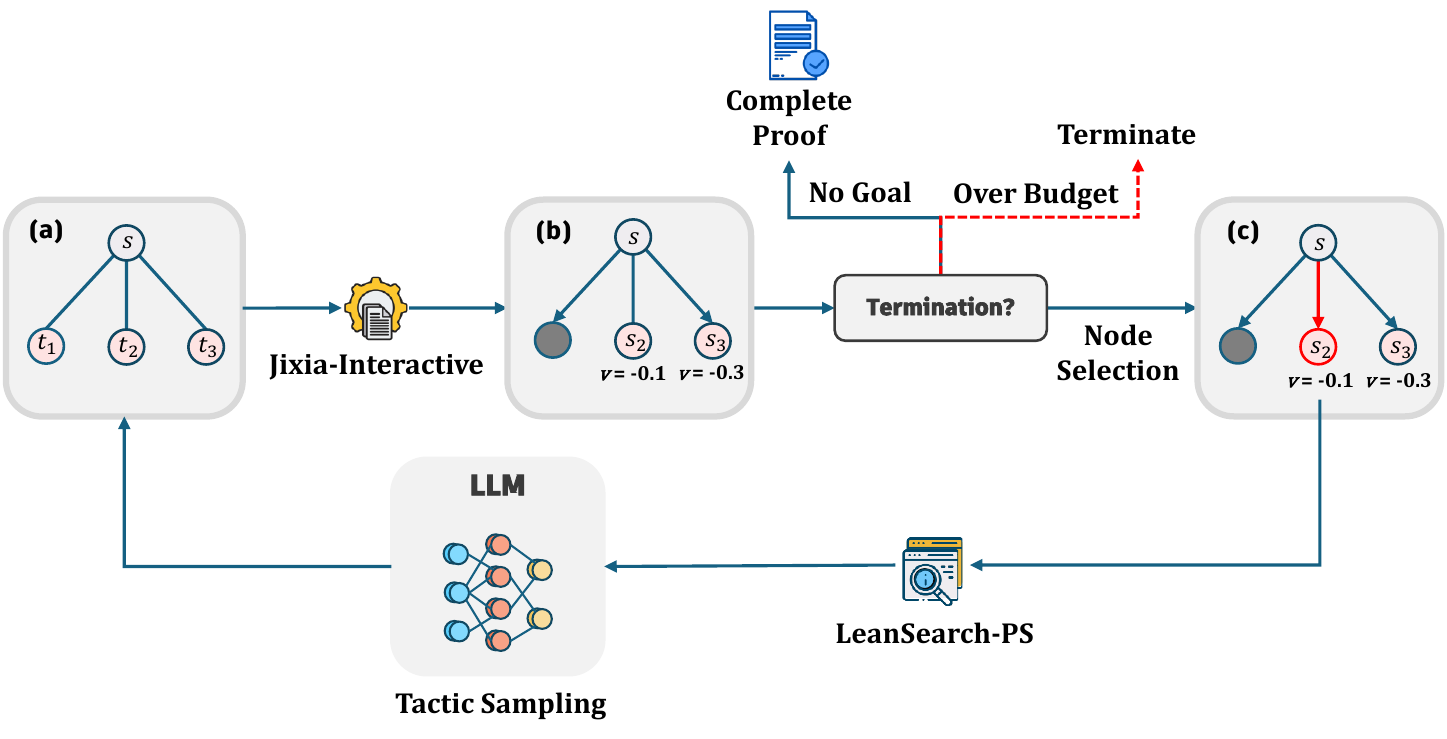}
        \caption{\textbf{Overview of REAL-Prover.} This figure illustrates three main processes: From (a) to (b), the prover passes several tactics to the Lean 4 interactive environment (Jixia-interactive) and receives corresponding scored state nodes. From (b) to (c), the prover selects a state node based on its score and checks whether the proof is complete. From (c) to (a), the prover using LLM with retrieval information from LeanSearch-PS generate new tactics.}
        \label{fig:reaper_fig}
\end{figure}

\subsection{REAL-Prover}
\label{subsec:real-prover}
From the HERALD-AF pipeline, we obtained a collection of formal statements without corresponding proofs. To generate complete proofs and achieve the second goal, we introduce the REAL-Prover system, illustrated in \Cref{fig:reaper_fig}. We developed a new Lean 4 interactive environment, Jixia-interactive, which receives Lean tactics and updates the proof state. For each formal statement, Jixia-interactive initializes the proof state. REAL-Prover constructs a search tree to explore possible proof paths. At each node, a LLM generates multiple candidate tactics. Jixia-interactive then applies these tactics, discards invalid ones, and advances to new proof states. Similar to BFS-Prover \cite{xin2025bfsproverscalablebestfirsttree}, each node is scored using the scoring function:$\frac{\sum_{t=0}^{L-1}\log p(a_t|s_t)}{L^{\alpha}}$,
where the $s_t$ represents the proof state and $a_t$ is the tactic applied at step t. $p(a_t|s_t)$ is the model’s predicted probability of applying tactic  $a_t$ at state $s_t$. $L$ is the total length of the path from the root to the current state $s_L$ and $\alpha$ is a hyper-parameter in $ [0,1]$.
We adopt a best-first search strategy, selecting the node with the highest score to expand further by generating new tactics. This iterative process continues until either the proof is completed or a predefined search budget is exhausted. Once completed, the full proof can be reconstructed from the search tree. To enhance our prover's performace, we add retrieval information from Mathlib using LeanSearch premise selection (LeanSearch-PS), which extends the work of LeanSearch \cite{gao2025semanticsearchenginemathlib4}. We will discuss the details of LeanSearch-PS.
\subsubsection{LeanSearch-PS}
In our work, premise selection addresses the challenge of identifying the most relevant theorems based on the current Lean proof state. To enhance this process, we developed LeanSearch-PS, a retrieval system built around a high-performance embedding model. This model projects both Lean proof states and Mathlib theorems into a shared semantic vector space. During proof generation, the current proof state is embedded, and a nearest-neighbor search is performed to retrieve the top-k most relevant theorems. These selected premises are then supplied as context to our language model, helping it produce more effective and accurate proof steps.

To train the embedding model, we construct a dataset of $(s, t_{\text{pos}})$ pairs extracted from Mathlib using the Jixia tool \cite{jixiaGitHub}. Here, $s$ denotes a Lean proof state, and $t_{\text{pos}}$ denotes the theorem successfully applied at that step. The training process adopts a two-stage framework inspired by transfer learning strategies commonly used in text retrieval \cite{SFRAIResearch2024}. In both stages, we employ the InfoNCE loss\cite{oord2018representation} defined as 
\begin{equation}
L(q,k_i)=-\log\frac{\exp(q\cdot k_0/\tau)}{\sum_{i=0}^n \exp(q\cdot k_i/\tau)},
\end{equation}
where $k_0$ is the embedding of positive example, $k_1, \ldots, k_n$ are the embedding of negative examples, and $\tau$ is a temperature hyperparameter. In the subsequent description, we will omit explicit mention of the embedding for brevity.

The training process consists of the following two stages.

\begin{itemize}[leftmargin = *]
\item \textbf{Initial Training:} We treat the proof state $s$ as the query $q$ and its corresponding theorem $t_{\text{pos}}$ as the positive example $k_0$. The in-batch negatives strategy is used, where other premises in the same batch serve as negative examples. This enables contrastive learning of premise relevance across proof states.

\item \textbf{Hard Negative Enhanced Training:} To refine fine-grained discrimination, we leverage the initially trained model to mine hard negatives—challenging but incorrect premises for each state. This yields triplets $(s, t_{\text{pos}}, t_{\text{hard\_neg}})$, where $t_{\text{hard\_neg}}$ denotes a mined hard negative premise. During this stage, we use $s$ as $q$, $t_{\text{pos}}$ as $k_0$, and $t_{\text{hard\_neg}}$ as $k_i$ $(i=1,\ldots,n)$. The resulting model after this phase constitutes our final embedding model.  
\end{itemize}

 \subsubsection{Expert Iteration}
  Another key component of REAL-Prover is the prover model within the tactic generator. To train a strong prover model, we combine expert iteration to collect synthetic data, as shown in \Cref{fig:overview_fig}(b). We begin by fine-tuning a prover model using open-source datasets. This initial model is then employed within REAL-Prover to solve formal statements produced by HERALD-AF, generating additional high-quality data. The newly collected data is incorporated into the supervised fine-tuning dataset to train an improved model. This refined model is then used to tackle challenging formal statements that the previous model could not solve.  We update our model continuously in this iteration process.

\section{Experiment Setup}
\subsection{Implementation Detail}
\label{sec:implementation_detail}

    
    
    

\paragraph{REAL-Prover Training}

We perform supervised fine-tuning on the base models Qwen2.5-Math-7B \cite{yang2024qwen2} using a learning rate of $5 \times 10^{-5}$ with a cosine decay scheduler and a maximum context length of 8192 tokens. Our training data comprises the following sources, totaling 210,420 state-tactic pairs:

\begin{enumerate}[label=\textbullet, leftmargin = *]
    \item Post-processed formal proofs collected via expert iteration. Each proof is parsed into state-tactic pairs using Jixia, and the corresponding prompts are enriched with natural language descriptions, the proof context, the current proof state, and retrieval information from LeanSearch-PS. The dataset comprises 15,071 algebra problems from undergraduate and graduate mathematics texts, and 19,633 NuminaMath problems \cite{numina_math_datasets}, resulting in 25,818 and 30,589 state-tactic pairs, respectively, after post-processing.
    
    \item Post-processed, human-annotated datasets were constructed following the same procedure as expert-iteration proofs. Based on 196 annotated undergraduate-level algebra problems, the datasets contain 18,669 state-tactic pairs. Annotations were provided by undergraduate and graduate students majoring in mathematics, ensuring high quality.

    \item Formal proofs are extracted from Mathlib and parsed into state-tactic pairs using Jixia. Prompt inputs are further augmented with retrieval information from LeanSearch-PS, while natural language descriptions are omitted. The resulting dataset comprises 92,152 state-tactic pairs.
    
    
    \item  The augmented Lean-Workbook dataset \cite{ying2024leanworkbooklargescalelean}. We enhance this dataset by adding retrieval information from LeanSearch-PS to each example, while keeping the original structure intact. The resulting dataset comprises 43,192 state-tactic pairs.
\end{enumerate}

\paragraph{LeanSearch-PS Training} We develop two dense retrievers from E5-mistral-7b-instruct \cite{wang2023improving} with Tevatron \cite{Gao2022TevatronAE} training framework. For both initial training and hard negative enhanced training, we use LoRA to accelerate training, and set the learning rate as $2 \times 10^{-5}$, the batch size as 128, the number of epoch as 1, and other hyperparameters as their default values. Queries are formal statements with a maximum length of 128 and passages are formal theorems with a maximum length of 256. For the hard negative examples, we first embed all statements and theorems with the initial trained embedding model, and then randomly select one passage from the top 30 to top 100 most similar ones for each query as its hard negative premise. 

\paragraph{REAL-Prover Inference}

In our experiments, considering the overall computational budget, we set the number of passes and samples per step to 64. We set our LLM's temperature to 1.5. Additionally, we set the hyperparameter $\alpha$ of the score function to 0.5, consistent with the configuration used in BFS-Prover \cite{xin2025bfsproverscalablebestfirsttree}
\subsection{Benchmark}
 As our prover highlights its capabilities on college-level mathematics, we created our own benchmark, FATE-M, which focuses specifically on algebraic formal statements to test model's performance on rich algebra structures. In addition, we also follow prior work \cite{xin2024deepseekprover,theoremllama,li2024hunyuanprover,lin2025goedel} to evaluate our prover's general capabilities using the ProofNet and MiniF2F benchmarks.

 \paragraph{ProofNet}ProofNet consists of 185 validation problems and 186 test problems, drawn from widely used undergraduate pure mathematics textbooks. The dataset spans a variety of topics, including real and complex analysis, linear algebra, abstract algebra, and topology.

\paragraph{MiniF2F} MiniF2F includes 244 validation problems and 244 test problems, collected from a diverse set of mathematical sources such as the AIME, AMC, and IMO competitions. This benchmark focuses on Olympiad-level problems covering core areas of elementary mathematics, including algebra, number theory, and mathematical induction.

\paragraph{FATE-M (Formal Algebra Theorem Evaluation-Medium)} FATE-M is a benchmark designed to evaluate theorem-proving capabilities in Lean4 for undergraduate-level abstract algebra. It consists of 141 problems formalized in Lean4, most accompanied by natural language comments in English or Chinese. Focusing on core topics in abstract algebra, FATE-M spans problems of simple to moderate difficulty, all of which rely on definitions and lemmas already formalized in Lean's Mathlib. To our knowledge, it is the first benchmark specifically targeting this domain, offering a specialized tool for assessing formal reasoning in algebra. Future expansions of the FATE series will include more advanced topics, such as commutative algebra, with increased difficulty.  

The benchmark was created by extracting problems from 12 university-level abstract algebra textbooks (for a full list, see \Cref{sec:textbook-sources}), which were then formalized by students. Each formalized statement was rigorously verified by Mathlib contributors and PhD students in algebra to ensure it accurately represents the original mathematical meaning. Additionally, Lean-checked proofs are provided for all statements to guarantee their correctness.  

We present two examples to illustrate the nature of the benchmark:

\begin{lstlisting}[language=Lean]
import Mathlib

example {G : Type*} [Group G] {A B : Subgroup G} : 
    (∃ C : Subgroup G, C.carrier = (A.carrier ∪ B.carrier)) ↔ A ≤ B ∨ B ≤ A := by
\end{lstlisting}

This is a medium-level problem in the benchmark. This problem translates to the following natural language statement: Let $A$, $B$, and $C$ be subgroups of a group $G$. If $C = A \cup B$, then either $A \subseteq B$ or $B \subseteq A$.
\newpage
\begin{lstlisting}[language=Lean]
import Mathlib

open Classical

/-Prove that a group of order $p^{2}, p$ a prime, has a normal subgroup of order $p$.-/
example {G : Type*} {p : ℕ} [Group G] [Fintype G] (pp : p.Prime) (ord : Fintype.card G = p ^ 2) : ∃ P : Subgroup G, (P.Normal) ∧ (Fintype.card P = p) := 
\end{lstlisting}

This problem represents one of the more challenging exercises in the benchmark. The natural language translation precisely matches the Lean code comment. An equivalent formulation would be: Let $p$ be a prime number. Then every group of order $p^2$ contains a normal subgroup of order $p$.

\section{Quantitative Result}
   Our main results are on ProofNet and FATE-M, both of which consist of college-level mathematics problems that require the use of numerous theorems. To evaluate the generalization ability of our system, we also test it on MiniF2F, a benchmark in a different domain than ProofNet and FATE-M. The results are shown below.
\subsection{ProofNet Result}

In the ProofNet benchmark, we compare REAL-Prover with several leading provers, Geodel-Prover \cite{lin2025goedel}, DeepSeek-Prover-V1.5 \cite{xin2024deepseekproverv15harnessingproofassistant}. For whole-proof system, the sampling budget in table means the number of generation. And for tree-search systems, the sampling budget is M$\times$N, where M is the total number of passes and N is the number of every step generation. The experimental results demonstrate that REAL-Prover with retrieval archieved 23.7\% success rate only using supervied fine-tune. Surpassing DeepSeek-Prover-V1.5-RL under a 3200-sampling budget and DeepSeek-Prover-V1.5-RL+RMaxTS under 1$\times$3200-sampling budget, which are trained with reinforcement learning.
\subsection{FATE-M Result}
\begin{table}[tbp]
    \caption{Comparison with state-of-the-art models (all 7B parameters) on ProofNet.}
    \centering
    \vskip 0.1in
    \begin{tabular}{lcc} 
        \toprule 
        Prove System & Sampling Budget & ProofNet Test \\
        \midrule 
        \multicolumn{3}{c}{\textbf{Whole-proof systems}} \\ 
        \midrule 
        Goedel Prover \cite{lin2025goedel} & 32 & 15.6 \% \\
        DeepSeek-Prover-V1.5-SFT  \cite{xin2024deepseekprover} & 128 & 15.9 \% \\
        DeepSeek-Prover-V1.5-RL  \cite{xin2024deepseekprover} & 128 & 18.2 \% \\
        \midrule 
        \multicolumn{3}{c}{\textbf{Tree search systems}} \\ 
        \midrule 
        DeepSeek-Prover-V1.5-RL + RMaxTS  \cite{xin2024deepseekproverv15harnessingproofassistant} & 1 × 3200 & 21.6 \% \\
        REAL-Prover (\textbf{ours})  & 64 × 64 & \textbf{23.7} \% \\
        
        \bottomrule 
    \end{tabular}

    \label{tab:proof-net}
\end{table}

The performance on the FATE-M benchmark is presented in \Cref{tab:fate-m}. Given the availability of open-source models and pipelines, we evaluate only Goedel-Prover and DeepSeek-Prover by using their official repositories. In this experiment, our prover achieves a state-of-the-art result with a Pass@64 of 56.7\%. This demonstrates that our prover exhibits strong capability in solving college-level algebraic problems.
\subsection{MiniF2F Result}

The MiniF2F benchmark has been the focus of more extensive research, as summarized in \Cref{tab:minif2f_result}. For tree search systems, serveral provers have K$\times$W$\times$N sampling budget, where K represents the number of passes, W the expansion width, and N the maximum number of proof state expansions per pass.  As shown in the table, our model performs relatively poorly compared to state-of-the-art models. We consider two main factors that may contribute to this performance gap. First, MiniF2F primarily features high school-level olympiad problems, which often do not heavily rely on theorems from Mathlib. Consequently, the benefit of retrieval is diminished compared to its effectiveness on college-level problems. Second, reinforcement learning plays a significant role in the training of current state-of-the-art provers. In contrast, our model is trained solely with supervised fine-tuning, making it challenging to compete without the additional boost provided by retrieval mechanisms.

\begin{table}[tbp]
    \centering
    \caption{Comparison with other system (all 7B parameters) on FATE-M.}
    \vskip 0.1in
    \begin{tabular}{lcc} 
        \toprule 
        Prove System & Sampling Budget & FATE-M Test \\
        \midrule 
        \multicolumn{3}{c}{\textbf{Whole-proof systems}} \\ 
        \midrule 
        Goedel Prover \cite{lin2025goedel}& 128 & 18.7 \% \\
        DeepSeek-Prover-V1.5-RL  \cite{xin2024deepseekprover} & 128 & 31.2 \% \\
        \midrule 
        \multicolumn{3}{c}{\textbf{Tree search systems}} \\ 
        \midrule 
        DeepSeek-Prover-V1.5-RL + RMaxTS \cite{xin2024deepseekproverv15harnessingproofassistant} & 64 × 64 & 41.8 \% \\
        REAL-Prover (\textbf{ours}) & 64 × 64 & \textbf{56.7} \% \\

        \bottomrule 
    \end{tabular}
    \label{tab:fate-m}
\end{table}
\begin{table}[tbp]
    \centering
    \caption{Comparison with other system on MiniF2F.} 
    \vskip 0.1in
    \begin{tabular}{lcc} 
        \toprule 
        Prove System & Sampling Budget & MiniF2F Test \\
        \midrule 
        \multicolumn{3}{c}{\textbf{Whole-proof systems}} \\ 
        \midrule 
        DeepSeek-Prover-V1.5-RL \cite{xin2024deepseekprover} & 102400 & 60.2\% \\
        Leanabell-Prover-GD-RL \cite{zhang2025leanabell} & 128 & 61.1\% \\
        Goedel-Prover \cite{lin2025goedel} & 25600 & 64.7\% \\
        STP \cite{dong2502stp} & 25600 & 67.6\% \\
        \midrule 
        \multicolumn{3}{c}{\textbf{Tree search systems}} \\ 
        \midrule 
        LLMStep \cite{welleck2023llmstep} & 1×32×100 & 27.9  \% \\
        GPT-f \cite{polu2020generativelanguagemodelingautomated} & 64×8×512 & 36.6  \% \\
        Hypertree Proof Search \cite{lample2022hypertree} & 64 × 5000 & 41.0 \% \\
        DeepSeek-Prover-V1.5-RL + RMaxTS  \cite{xin2024deepseekproverv15harnessingproofassistant} & 32 × 6400 & 63.5 \% \\
        InternLM2.5-StepProver-BF \cite{wu2024internlm2} & 256 × 32 × 600 & 65.9 \% \\
        HunyuanProver v16 + BFS + DC \cite{li2024hunyuanprover} & 600 × 8 × 400 & 68.4 \% \\
        BFS-Prover \cite{xin2025bfsproverscalablebestfirsttree} & 2048 × 2 × 600 & 70.8 \% \\
        REAL-Prover (\textbf{ours}) & 64 × 64 & 54.1 \% \\
 
        \bottomrule 
    \end{tabular}
    \label{tab:minif2f_result}
\end{table}

\subsection{Ablation Study}
\begin{table}[tbp]
    \centering
    \caption{Ablation study on ProofNet Test and FATE-M Test. REAL-Prover-v1-NoRet w/o LeanSearch refers to use the REAL-Prover-v1-NoRet model without LeanSearch-PS. REAL-Prover-v1 w/ LeanSearch refers to use the REAL-Prover-v1 model with LeanSearch-PS enabled.} 
    \vskip 0.1in
    \begin{tabular}{lcc} 
        \toprule 
        Prove System & ProofNet Test & FATE-M Test \\
        \midrule 
        REAL-Prover-v1-NoRet w/o LeanSearch & 22.6\% & 44.7\% \\
        REAL-Prover-v1 w/  LeanSearch & 23.7\% & 56.7\% \\
        \bottomrule 
    \end{tabular}
    \label{tab:ablation_study}
\end{table}
In this section, we will analysis the effective of LeanSearch-PS. To compare with REAL-Prover-v1, we trained a prover model, REAL-Prover-v1-NoRet, which used the same dataset but without retrieval information. And in REAL-Prover, we do not use LeanSearch-PS for REAL-Prover-v1-NoRet model. We evaluate both models on the ProofNet and FATE-M benchmarks with results presented in \Cref{tab:ablation_study}.  The results show that REAL-Prover-v1 achieves a performance improvement over REAL-Prover-v1-NoRet on both benchmarks. This demonstrates that the retrieval system enhances the prover's performance on college-level mathematics problems.
 
\begin{figure}[!htbp]
    \centering
    \includegraphics[width=\textwidth]{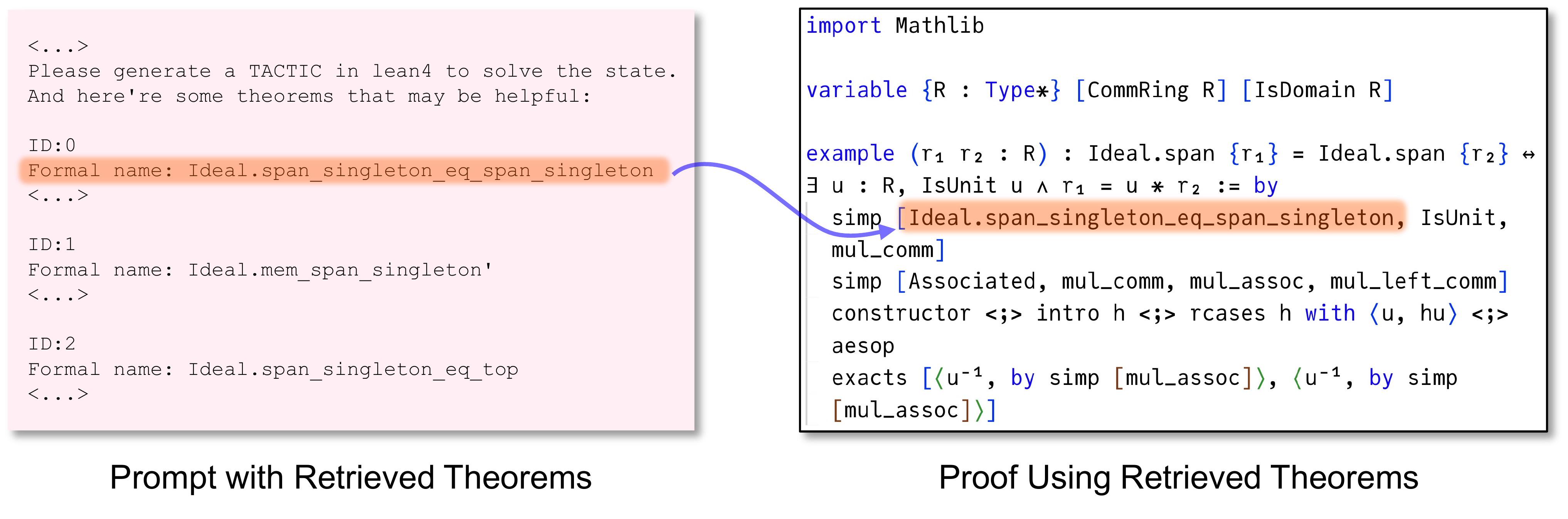}
    \caption{Proof using LeanSearch-PS and retrieval results. The left shows part of our input prompt; the right displays the proof generated by REAL-Prover.}
    \label{fig:retrieval_result}
\end{figure}

We present a comparison between proofs with and without LeanSearch-PS. In \Cref{fig:retrieval_result}, the proof assisted by LeanSearch-PS successfully retrieves  critical lemmas and completes verification, whereas the proof without LeanSearch-PS fails to verify. For problems that can be resolved both with LeanSearch-PS and without LeanSearch-PS, the one with LeanSearch-PS often leads to more accurate and human-readable proof, we provide 2 examples in \Cref{case_study} to illustrate this observation.


\section{Conclusion and Future Work}
In this work, we propose REAL-Prover, a stepwise proof system designed to tackle challenging mathematical problems. To support its development, we introduce the HERALD-AF pipeline, which translates informal mathematical statements into formal Lean~4 representations, enabling the construction of a large-scale formal dataset. We also develop a new interactive Lean~4 environment, Jixia-Interactive, which facilitates both the training of our prover model, REAL-Prover-v1, and efficient formal proof generation. Furthermore, we present LeanSearch-PS, a theorem retrieval system that improves theorem selection and boosts overall proof success on college-level mathematical benchmarks. To comprehensively evaluate the prover’s effectiveness, we propose the FATE-M benchmark, on which REAL-Prover achieves a state-of-the-art success rate of 56.7\%.

Despite these promising results, several limitations remain. The current model is trained purely with supervised objectives, and we have not yet incorporated reinforcement learning (RL) due to its computational demands.  Nevertheless, prior studies suggest that RL can unlock latent reasoning capabilities in large language models, indicating that RL-based fine-tuning (e.g., GRPO~\cite{shao2024deepseekmath}) constitutes a promising avenue for enhancing performance. In addition, REAL-Prover emits only the final Lean tactics, whereas several leading provers first generate a natural-language chain of thought (CoT) and then translate it into formal steps, enabling powerful test-time reasoning amplification. Introducing an informal-reasoning layer would let the model draft, filter, and organize candidate arguments before committing to formal tactics, potentially improving both premise selection and high-level proof planning.

Future work will focus on these two complementary directions: incorporating RL-based fine-tuning to deepen reasoning capabilities and generalization, and integrating CoT-driven informal reasoning to bridge natural and formal mathematical discourse. These extensions are expected to yield a more autonomous, interpretable, and robust formal reasoning system, advancing the broader objective of scalable, human-aligned mathematical intelligence.

\section*{Acknowledgements}
This work is supported in part by National Key R\&D Program of China grant 2024YFA1014000, the New Cornerstone Investigator Program, and Ubiquant.
\small
\bibliographystyle{unsrt}
\bibliography{references}

@inproceedings{theoremllama,
    title = "{T}heorem{L}lama: Transforming General-Purpose {LLM}s into Lean4 Experts",
    author = "Wang, Ruida  and
      Zhang, Jipeng  and
      Jia, Yizhen  and
      Pan, Rui  and
      Diao, Shizhe  and
      Pi, Renjie  and
      Zhang, Tong",
    editor = "Al-Onaizan, Yaser  and
      Bansal, Mohit  and
      Chen, Yun-Nung",
    booktitle = "Proceedings of the 2024 Conference on Empirical Methods in Natural Language Processing",
    month = nov,
    year = "2024",
    address = "Miami, Florida, USA",
    publisher = "Association for Computational Linguistics",
    url = "https://aclanthology.org/2024.emnlp-main.667/",
    doi = "10.18653/v1/2024.emnlp-main.667",
    pages = "11953--11974"
}

@inproceedings{
xin2024deepseekprover,
title={Advancing Theorem Proving in {LLM}s through Large-Scale Synthetic Data},
author={Huajian Xin and Daya Guo and Zhihong Shao and Z.Z. Ren and Qihao Zhu and Bo Liu and Chong Ruan and Wenda Li and Xiaodan Liang},
booktitle={The 4th Workshop on Mathematical Reasoning and AI at NeurIPS'24},
year={2024},
url={https://openreview.net/forum?id=TPtXLihkny}
}

@inproceedings{
kumarappan2024leanagent,
title={LeanAgent: Lifelong Learning for Formal Theorem Proving},
author={Adarsh Kumarappan and Mo Tiwari and Peiyang Song and Robert Joseph George and Chaowei Xiao and Anima Anandkumar},
booktitle={The Thirteenth International Conference on Learning Representations},
year={2025},
url={https://openreview.net/forum?id=Uo4EHT4ZZ8}
}

@inproceedings{karpukhin2020dense,
  title={Dense Passage Retrieval for Open-Domain Question Answering.},
  author={Karpukhin, Vladimir and Oguz, Barlas and Min, Sewon and Lewis, Patrick SH and Wu, Ledell and Edunov, Sergey and Chen, Danqi and Yih, Wen-tau},
  booktitle={EMNLP (1)},
  pages={6769--6781},
  year={2020}
}

@article{miniF2F,
  author       = {Kunhao Zheng and
                  Jesse Michael Han and
                  Stanislas Polu},
  title        = {MiniF2F: a cross-system benchmark for formal Olympiad-level mathematics},
  journal      = {CoRR},
  volume       = {abs/2109.00110},
  year         = {2021},
  url          = {https://arxiv.org/abs/2109.00110},
  eprinttype    = {arXiv},
  eprint       = {2109.00110},
  bibsource    = {dblp computer science bibliography, https://dblp.org}
}

@inproceedings{
gao2025heraldnaturallanguageannotated,
title={Herald: A Natural Language Annotated Lean 4 Dataset},
author={Guoxiong Gao and Yutong Wang and Jiedong Jiang and Qi Gao and Zihan Qin and Tianyi Xu and Bin Dong},
booktitle={The Thirteenth International Conference on Learning Representations},
year={2025},
url={https://openreview.net/forum?id=Se6MgCtRhz}
}

@inproceedings{
xin2024deepseekproverv15harnessingproofassistant,
title={DeepSeek-Prover-V1.5: Harnessing Proof Assistant Feedback for Reinforcement Learning and Monte-Carlo Tree Search},
author={Huajian Xin and Z.Z. Ren and Junxiao Song and Zhihong Shao and Wanjia Zhao and Haocheng Wang and Bo Liu and Liyue Zhang and Xuan Lu and Qiushi Du and Wenjun Gao and Haowei Zhang and Qihao Zhu and Dejian Yang and Zhibin Gou and Z.F. Wu and Fuli Luo and Chong Ruan},
booktitle={The Thirteenth International Conference on Learning Representations},
year={2025},
url={https://openreview.net/forum?id=I4YAIwrsXa}
}

@inproceedings{gao2025semanticsearchenginemathlib4,
    title = "A Semantic Search Engine for Mathlib4",
    author = "Gao, Guoxiong  and
      Ju, Haocheng  and
      Jiang, Jiedong  and
      Qin, Zihan  and
      Dong, Bin",
    editor = "Al-Onaizan, Yaser  and
      Bansal, Mohit  and
      Chen, Yun-Nung",
    booktitle = "Findings of the Association for Computational Linguistics: EMNLP 2024",
    month = nov,
    year = "2024",
    address = "Miami, Florida, USA",
    publisher = "Association for Computational Linguistics",
    url = "https://aclanthology.org/2024.findings-emnlp.470/",
    doi = "10.18653/v1/2024.findings-emnlp.470",
    pages = "8001--8013"
}

@article{shao2024deepseekmath,
  author = {Zhihong Shao, Peiyi Wang, Qihao Zhu and others},
  title = {DeepSeekMath: Pushing the Limits of Mathematical Reasoning in Open Language Models},
  journal = {CoRR},
  volume = {abs/2402.03300},
  year = {2024},
  url = {https://arxiv.org/abs/2402.03300},
}

@inproceedings{
xin2024deepseek,
title={Advancing Theorem Proving in {LLM}s through Large-Scale Synthetic Data},
author={Huajian Xin and Daya Guo and Zhihong Shao and Z.Z. Ren and Qihao Zhu and Bo Liu and Chong Ruan and Wenda Li and Xiaodan Liang},
booktitle={The 4th Workshop on Mathematical Reasoning and AI at NeurIPS'24},
year={2024},
url={https://openreview.net/forum?id=TPtXLihkny}
}

@article{xin2025bfsproverscalablebestfirsttree,
      title={BFS-Prover: Scalable Best-First Tree Search for LLM-based Automatic Theorem Proving}, 
      author={Ran Xin and Chenguang Xi and Jie Yang and Feng Chen and Hang Wu and Xia Xiao and Yifan Sun and Shen Zheng and Kai Shen},
      year={2025},
      eprint={2502.03438},
      archivePrefix={arXiv},
      primaryClass={cs.AI},
journal={arXiv preprint arXiv:2502.03438},  
      url={https://arxiv.org/abs/2502.03438}, 
}

@article{azerbayev2023proofnetautoformalizingformallyproving,
      title={ProofNet: Autoformalizing and Formally Proving Undergraduate-Level Mathematics}, 
      author={Zhangir Azerbayev and Bartosz Piotrowski and Hailey Schoelkopf and Edward W. Ayers and Dragomir Radev and Jeremy Avigad},
journal={arXiv preprint arXiv:2302.12433},
      year={2023},
      eprint={2302.12433},
      archivePrefix={arXiv},
      primaryClass={cs.CL},
      url={https://arxiv.org/abs/2302.12433}, 
}

@inproceedings{
lample2022hypertree,
title={HyperTree Proof Search for Neural Theorem Proving},
author={Guillaume Lample and Timothee Lacroix and Marie-anne Lachaux and Aurelien Rodriguez and Amaury Hayat and Thibaut Lavril and Gabriel Ebner and Xavier Martinet},
booktitle={Advances in Neural Information Processing Systems},
editor={Alice H. Oh and Alekh Agarwal and Danielle Belgrave and Kyunghyun Cho},
year={2022},
url={https://openreview.net/forum?id=J4pX8Q8cxHH}
}

@article{dong2502stp,
      title={STP: Self-play LLM Theorem Provers with Iterative Conjecturing and Proving}, 
      author={Kefan Dong and Tengyu Ma},
    journal={arXiv preprint arXiv:2502.00212},
      year={2025},
      eprint={2502.00212},
      archivePrefix={arXiv},
      primaryClass={cs.LG},
      url={https://arxiv.org/abs/2502.00212}, 
}

@article{li2024hunyuanprover,
  title={Hunyuanprover: A scalable data synthesis framework and guided tree search for automated theorem proving},
  author={Li, Yang and Du, Dong and Song, Linfeng and Li, Chen and Wang, Weikang and Yang, Tao and Mi, Haitao},
  journal={arXiv preprint arXiv:2412.20735},
  year={2024}
}

@article{lin2025goedel,
      title={Goedel-Prover: A Frontier Model for Open-Source Automated Theorem Proving}, 
      author={Yong Lin and Shange Tang and Bohan Lyu and Jiayun Wu and Hongzhou Lin and Kaiyu Yang and Jia Li and Mengzhou Xia and Danqi Chen and Sanjeev Arora and Chi Jin},
      year={2025},
      eprint={2502.07640},
      archivePrefix={arXiv},
      primaryClass={cs.LG},
      url={https://arxiv.org/abs/2502.07640}, 
      journal={arXiv preprint arXiv:2502.07640},  
}

@article{wu2024internlm2,
      title={InternLM2.5-StepProver: Advancing Automated Theorem Proving via Expert Iteration on Large-Scale LEAN Problems}, 
      author={Zijian Wu and Suozhi Huang and Zhejian Zhou and Huaiyuan Ying and Jiayu Wang and Dahua Lin and Kai Chen},
      year={2024},
      eprint={2410.15700},
      archivePrefix={arXiv},
      primaryClass={cs.AI},
      journal={arXiv preprint arXiv:2410.15700}
}

@article{zhang2025leanabell,
  title={Leanabell-prover: Posttraining scaling in formal reasoning},
  author={Zhang, Jingyuan and Wang, Qi and Ji, Xingguang and Liu, Yahui and Yue, Yang and Zhang, Fuzheng and Zhang, Di and Zhou, Guorui and Gai, Kun},
  journal={arXiv preprint arXiv:2504.06122},
  year={2025}
}

@inproceedings{wang2023improving,
    title = "Improving Text Embeddings with Large Language Models",
    author = "Wang, Liang  and
      Yang, Nan  and
      Huang, Xiaolong  and
      Yang, Linjun  and
      Majumder, Rangan  and
      Wei, Furu",
    editor = "Ku, Lun-Wei  and
      Martins, Andre  and
      Srikumar, Vivek",
    booktitle = "Proceedings of the 62nd Annual Meeting of the Association for Computational Linguistics (Volume 1: Long Papers)",
    month = aug,
    year = "2024",
    address = "Bangkok, Thailand",
    publisher = "Association for Computational Linguistics",
    url = "https://aclanthology.org/2024.acl-long.642/",
    doi = "10.18653/v1/2024.acl-long.642",
    pages = "11897--11916"
}

@inproceedings{Gao2022TevatronAE,
author = {Gao, Luyu and Ma, Xueguang and Lin, Jimmy and Callan, Jamie},
title = {Tevatron: An Efficient and Flexible Toolkit for Neural Retrieval},
year = {2023},
isbn = {9781450394086},
publisher = {Association for Computing Machinery},
address = {New York, NY, USA},
url = {https://doi.org/10.1145/3539618.3591805},
doi = {10.1145/3539618.3591805},
booktitle = {Proceedings of the 46th International ACM SIGIR Conference on Research and Development in Information Retrieval},
pages = {3120–3124},
numpages = {5},
location = {Taipei, Taiwan},
series = {SIGIR '23}
}

@misc{jixiaGitHub,
  author       = {frenzymath},
  title        = {jixia: A static analysis tool for Lean 4},
  howpublished = {\url{https://github.com/frenzymath/jixia}},
  year         = {2024}
}

@misc{SFRAIResearch2024,
  title={{SFR-Embedding-Mistral}: Enhance Text Retrieval with Transfer Learning},
  author={Rui Meng and Ye Liu and Shafiq Rayhan Joty and Caiming Xiong and Yingbo Zhou and Semih Yavuz},
  howpublished={Salesforce AI Research Blog},
  year={2024}
}

@InProceedings{fourcolourthm,
author="Gonthier, Georges",
editor="Kapur, Deepak",
title="The Four Colour Theorem: Engineering of a Formal Proof",
booktitle="Computer Mathematics",
year="2008",
publisher="Springer Berlin Heidelberg",
address="Berlin, Heidelberg",
pages="333--333",
isbn="978-3-540-87827-8"
}

@InProceedings{oddthm,
author="Gonthier, Georges
and Asperti, Andrea
and Avigad, Jeremy
and Bertot, Yves
and Cohen, Cyril
and Garillot, Fran{\c{c}}ois
and Le Roux, St{\'e}phane
and Mahboubi, Assia
and O'Connor, Russell
and Ould Biha, Sidi
and Pasca, Ioana
and Rideau, Laurence
and Solovyev, Alexey
and Tassi, Enrico
and Th{\'e}ry, Laurent",
editor="Blazy, Sandrine
and Paulin-Mohring, Christine
and Pichardie, David",
title="A Machine-Checked Proof of the Odd Order Theorem",
booktitle="Interactive Theorem Proving",
year="2013",
publisher="Springer Berlin Heidelberg",
address="Berlin, Heidelberg",
pages="163--179",
isbn="978-3-642-39634-2"
}

@inproceedings{
ying2024leanworkbooklargescalelean,
title={Lean Workbook: A large-scale Lean problem set formalized from natural language math problems},
author={Huaiyuan Ying and Zijian Wu and Yihan Geng and JIayu Wang and Dahua Lin and Kai Chen},
booktitle={The Thirty-eight Conference on Neural Information Processing Systems Datasets and Benchmarks Track},
year={2024},
url={https://openreview.net/forum?id=Vcw3vzjHDb}
}

@article{ren2025deepseekproverv2advancingformalmathematical,
      title={DeepSeek-Prover-V2: Advancing Formal Mathematical Reasoning via Reinforcement Learning for Subgoal Decomposition}, 
      author={Z. Z. Ren and Zhihong Shao and Junxiao Song and Huajian Xin and Haocheng Wang and Wanjia Zhao and Liyue Zhang and Zhe Fu and Qihao Zhu and Dejian Yang and Z. F. Wu and Zhibin Gou and Shirong Ma and Hongxuan Tang and Yuxuan Liu and Wenjun Gao and Daya Guo and Chong Ruan},
journal={arXiv preprint arXiv:2504.21801},
      year={2025},
      eprint={2504.21801},
      archivePrefix={arXiv},
      primaryClass={cs.CL},
      url={https://arxiv.org/abs/2504.21801}, 
}

@InProceedings{lean,
author="Moura, Leonardo de
and Ullrich, Sebastian",
editor="Platzer, Andr{\'e}
and Sutcliffe, Geoff",
title="The Lean 4 Theorem Prover and Programming Language",
booktitle="Automated Deduction -- CADE 28",
year="2021",
publisher="Springer International Publishing",
address="Cham",
pages="625--635",
isbn="978-3-030-79876-5"
}

@book{coq,
  title={Interactive theorem proving and program development: Coq’Art: the calculus of inductive constructions},
  author={Bertot, Yves and Cast{\'e}ran, Pierre},
  year={2013},
  publisher={Springer Science \& Business Media}
}

@inproceedings{
yang2023leandojotheoremprovingretrievalaugmented,
title={LeanDojo: Theorem Proving with Retrieval-Augmented Language Models},
author={Kaiyu Yang and Aidan M Swope and Alex Gu and Rahul Chalamala and Peiyang Song and Shixing Yu and Saad Godil and Ryan Prenger and Anima Anandkumar},
booktitle={Thirty-seventh Conference on Neural Information Processing Systems Datasets and Benchmarks Track},
year={2023},
url={https://openreview.net/forum?id=g7OX2sOJtn}
}

@inproceedings{The_mathlib_Community_2020, series={POPL ’20},
   title={The lean mathematical library},
   url={http://dx.doi.org/10.1145/3372885.3373824},
   DOI={10.1145/3372885.3373824},
   booktitle={Proceedings of the 9th ACM SIGPLAN International Conference on Certified Programs and Proofs},
   publisher={ACM},
   author={The mathlib Community},
   year={2020},
   month=jan, collection={POPL ’20} 
}

@inproceedings{
jiang2023draftsketchproveguiding,
title={Draft, Sketch, and Prove: Guiding Formal Theorem Provers with Informal Proofs},
author={Albert Qiaochu Jiang and Sean Welleck and Jin Peng Zhou and Timothee Lacroix and Jiacheng Liu and Wenda Li and Mateja Jamnik and Guillaume Lample and Yuhuai Wu},
booktitle={The Eleventh International Conference on Learning Representations },
year={2023},
url={https://openreview.net/forum?id=SMa9EAovKMC}
}

@inproceedings{
zhao2023decomposingenigmasubgoalbaseddemonstration,
title={Subgoal-based Demonstration Learning for Formal Theorem Proving},
author={Xueliang Zhao and Wenda Li and Lingpeng Kong},
booktitle={Forty-first International Conference on Machine Learning},
year={2024},
url={https://openreview.net/forum?id=pSnhA7Em1P}
}

@inproceedings{
wang2023legoproverneuraltheoremproving,
title={{LEGO}-Prover: Neural Theorem Proving with Growing Libraries},
author={Haiming Wang and Huajian Xin and Chuanyang Zheng and Zhengying Liu and Qingxing Cao and Yinya Huang and Jing Xiong and Han Shi and Enze Xie and Jian Yin and Zhenguo Li and Xiaodan Liang},
booktitle={The Twelfth International Conference on Learning Representations},
year={2024},
url={https://openreview.net/forum?id=3f5PALef5B}
}

@article{welleck2023llmstep,
  title={Llmstep: Llm proofstep suggestions in lean},
  author={Welleck, Sean and Saha, Rahul},
  journal={arXiv preprint arXiv:2310.18457},
  year={2023}
}

@article{polu2020generativelanguagemodelingautomated,
      title={Generative Language Modeling for Automated Theorem Proving}, 
      author={Stanislas Polu and Ilya Sutskever},
      journal={arXiv preprint arXiv:2009.03393},
      year={2020},
      eprint={2009.03393},
      archivePrefix={arXiv},
      primaryClass={cs.LG},
      url={https://arxiv.org/abs/2009.03393}, 
}

@article{deepseekai2025deepseekv3technicalreport,
      title={DeepSeek-V3 Technical Report}, 
      author={DeepSeek-AI and Aixin Liu and Bei Feng and Bing Xue and Bingxuan Wang and others},
journal={arXiv preprint arXiv:2412.19437},
      year={2025},
      eprint={2412.19437},
      archivePrefix={arXiv},
      primaryClass={cs.CL},
      url={https://arxiv.org/abs/2412.19437}, 
}

@article{yang2024qwen2,
  title={Qwen2 technical report},
  author={Yang, An and Yang, Baosong and Hui, Binyuan and Zheng, Bo and Yu, Bowen and Zhou, Chang and Li, Chengpeng and Li, Chengyuan and Liu, Dayiheng and Huang, Fei and others},
  journal={arXiv preprint arXiv:2407.10671},
  year={2024}
}

@article{numina_math_datasets,
  author = {Jia LI and Edward Beeching and Lewis Tunstall and Ben Lipkin and Roman Soletskyi and Shengyi Costa Huang and Kashif Rasul and Longhui Yu and Albert Jiang and Ziju Shen and Zihan Qin and Bin Dong and Li Zhou and Yann Fleureau and Guillaume Lample and Stanislas Polu},
  title = {NuminaMath},
  year = {2024},
  publisher = {Numina},
  journal = {Hugging Face repository},
  howpublished = {\url{[https://huggingface.co/AI-MO/NuminaMath-CoT](https://github.com/project-numina/aimo-progress-prize/blob/main/report/numina_dataset.pdf)}}
}

@article{oord2018representation,
  title={Representation learning with contrastive predictive coding},
  author={Oord, Aaron van den and Li, Yazhe and Vinyals, Oriol},
  journal={arXiv preprint arXiv:1807.03748},
  year={2018}
}
\appendix
\section{Prompt for REAL-Prover}
\tcbset{
  mybox/.style={
    colback=gray!10,      
    colframe=gray!100,     
    arc=4mm,              
    boxrule=0.8pt,        
    left=4pt, right=4pt,  
    top=4pt, bottom=4pt,
    enhanced,
  }
}

\begin{tcolorbox}[mybox]
In Lean, a formal proof is a fully constructed proof term that is type-checked and verified by the kernel. It represents a complete and correct derivation of a proposition.

The state after tactics refers to the intermediate proof state during tactic-based proof construction. It includes the list of remaining goals and the local context at that point.   
Relationship:
\begin{enumerate}
    \item Tactics are procedural tools used to incrementally construct a formal proof.
    \item Each tactic transforms the current proof state by solving or reducing goals.
    \item The state after a tactic reflects the goals that still need to be proven after that tactic has been applied.
    \item Once all goals are solved, Lean assembles the underlying proof terms generated by the tactics into a complete formal proof.
    \item This final term is then type-checked by the kernel to ensure correctness.
\end{enumerate}

In essence, the state after tactics shows where you are in the process of building a formal proof — it's a snapshot of what's left to do before the proof is complete.

Here is the FORMAL PROOF before the current state:
\begin{lstlisting}[language=Lean]
import Mathlib
open BigOperators
open Real Nat Topology

theorem parallelogram_midpoints_coincide_extracted {A B C D : ℂ} : (A + B + C + D) / 4 = (A + C) / 2 → (B + D) / 2 = (A + C) / 2 → B.re + D.re = 2 * ((A + C) / 2).re  := by
simp [add_comm, add_left_comm, add_assoc]
simp [div_eq_mul_inv, mul_add, mul_comm, mul_left_comm]
intro h₁ h₂

\end{lstlisting}

Here is the current STATE:
\begin{lstlisting}[language=Lean]
A : ℂ
B : ℂ
C : ℂ
D : ℂ
h₁ : (A + (B + (C + D))) * 4⁻¹ = (A + C) * 2⁻¹
h₂ : B + D = A + C
⊢ B.re + D.re = A.re + C.re
\end{lstlisting}
Please generate a tactic in lean4 to solve the state.

And here're some theorems that may be helpful:

ID:0

Formal name:\lstinline[language=Lean]|Complex.add_re|

Informal name: Real Part of Sum of Complex Numbers

Formal statement: \lstinline[language=Lean]|∀ (z w : ℂ), (z + w).re = z.re + w.re|





...



ID:5

Formal name: \lstinline[language=Lean]|add_add_add_comm|

Informal name: Commutativity and Associativity of Addition in Additive Commutative Semigroups

Formal statement: \lstinline[language=Lean]|∀ {G : Type u_3} [inst : AddCommSemigroup G] (a b c d : G), a + b + (c + d) = a + c + (b + d)|

\end{tcolorbox}

\newpage

\section{Source Textbooks for FATE-M}
\label{sec:textbook-sources}

The FATE-M benchmark draws problems from the following abstract algebra textbooks:

\begin{enumerate}[leftmargin = *]
    \item Pinter, C. C. \textit{A Book of Abstract Algebra} (2nd ed., 2010)
    \item Fraleigh, J. B. \textit{A First Course in Abstract Algebra} (8th ed., 2021)
    \item Anderson, M. \& Feil, T. \textit{A First Course in Abstract Algebra: Rings, Groups, and Fields} (3rd ed., 2015)
    \item Rotman, J. J. \textit{A First Course in Abstract Algebra with Applications} (3rd ed., 2005)
    \item Hodge, J. K., Schlicker, S., \& Sundstrom, T. \textit{Abstract Algebra: An Inquiry-Based Approach} (2014)
    \item Herstein, I. N. \textit{Abstract Algebra} (3rd ed., 1996)
    \item Grillet, P. A. \textit{Abstract Algebra} (2nd ed., 2007)
    \item Choudhary, P. \textit{Abstract Algebra} (2008)
    \item Saracino, D. \textit{Abstract Algebra: A First Course} (2nd ed., 2008)
    \item Hungerford, T. W. \textit{Abstract Algebra: An Introduction} (3rd ed., 2014)
    \item Dummit, D. S. \& Foote, R. M. \textit{Abstract Algebra} (3rd ed., 2004)
    \item Feng, K. \& Zhang, P. \textit{300 Problems in Abstract Algebra} (2009)
\end{enumerate}

\section{Training Details}

\paragraph{REAL-Prover training} REAL-Prover training is performed on 8$\times$A800 GPUs (80GB each) for approximately 18 hours, with hyperparameters detailed in \Cref{hyperparameters}.

\begin{table}[tbp]
    \centering
    \vskip 0.1in
    \caption{REAL-Prover training  hyperparameters}
    \begin{tabular}{ll}
    \toprule
    \textbf{Component}      & \textbf{Setting} \\
    \midrule
    Full Fine Tuning        &  Learning rate : $5\times 10^{-5}$; Scheducler: Cosine Decay \\
    \midrule
    Backbone                & Qwen2.5-Math-7B \\
    \midrule
    Length                  & Prompt Max Length: 8192\\
    \midrule
    Optimizations           & \lstinline|bf16, flash_attn|\\
    \midrule
    Batch Sizes             & \lstinline|train_batch_size = 2| \\
    \midrule
    Training Schedule       & 3 Epochs\\
    \bottomrule
    \end{tabular}
    \label{hyperparameters}
\end{table}

\paragraph{LeanSearch-PS training} LeanSearch-PS training is performed on 4$\times$L40 GPUs (40GB each) for approximately 12 hours with hyperparameters detailed in \Cref{hyperparameters2}.

\begin{table}[tbp]
    \centering
    \vskip 0.1in
    \caption{LeanSearch-PS training hyperparameters}
    \begin{tabular}{ll}
    \toprule
    \textbf{Component}      & \textbf{Setting} \\
    \midrule
    LoRA Fine Tuning        &  Learning rate : $2\times 10^{-5}$ \\
    \midrule
    Backbone                & E5-mistral-7b-instruct \\
    \midrule
    Optimizations           & \lstinline|bf16, flash_attn|\\
    \midrule
    Batch Sizes             & \lstinline|train_batch_size = 2| \\
    \midrule
    Training Schedule       & 1 Epoch\\
    \bottomrule
    \end{tabular}
    \label{hyperparameters2}
\end{table}

\section{Broader Impacts}
This work explores the use of large language models (LLMs) for automated formal theorem proving. Potential positive impacts include:
\begin{enumerate}
    \item assisting mathematicians in constructing formal proofs more efficiently by automating tedious or routine steps;
    \item facilitating the teaching of formal proof assistants such as Lean in undergraduate and graduate curricula;
    \item potentially enabling LLMs to autonomously explore new conjectures or discover novel proofs, contributing to mathematical discovery.
\end{enumerate}

However, several risks and limitations should be acknowledged. First, while LLMs may generate correct formal proofs, these proofs can often be opaque or unintuitive, reducing their explanatory value for human users. Second, excessive reliance on automated proof generation may hinder the development of deep mathematical understanding, particularly in educational settings. Third, the reinforcement learning and training process may be susceptible to reward hacking, especially given the complexity and non-determinism of the Lean environment—raising the risk of generating formally valid but semantically meaningless or misleading proofs that may be overlooked by human reviewers.

Careful evaluation, interpretability mechanisms, and human-in-the-loop verification are essential to ensure that such systems are used responsibly and contribute positively to both research and education.

\section{Case study}\label{case_study}

We provide 2 cases in FATE-M to illustrate that even if both REAL-Prover-v1 with LeanSearch-PS and REAL-Prover-v1-NoRet without Leansearch-PS produce the valid results, the one with premise selection often produce faster and readable proofs, as illustrated in \Cref{fig:ragnrag1} and \Cref{fig:ragnrag2}.

\begin{figure}[tbp]
    \begin{subfigure}{0.45\textwidth}
    \vtop{
        \begin{lstlisting}
example  {F : Type*} [Field F] [Fintype F] [IsAlgClosed F] : False := by
  set e : Fintype F := Fintype.ofFinite F
  set L := (AlgebraicClosure F)
  have : Infinite L := by exact IsAlgClosed.instInfinite
  have := this
  have := Fintype.ofFinite F
  exact Fintype.false e
        \end{lstlisting}        
        \caption{Proof with LeanSearch-PS.}
        \label{fig:with_leansearch}
    }
    \end{subfigure}
    \hspace{0.02\textwidth}
    \begin{subfigure}{0.48\textwidth}
    \vtop{
        \begin{lstlisting}        
example  {F : Type*} [Field F] [Fintype F] [IsAlgClosed F] : False := by
  have : Fact (Nat.Prime 2) := ⟨by norm_num⟩
  exact not_finite F
        \end{lstlisting}        
        \caption{Proof without LeanSearch-PS.}
        \label{fig:without_leansearch}
    }
    \end{subfigure}
\caption{Compare the proofs with and without LeanSearch-PS. The prover in (a) uses the existing instance `IsAlgClosed.instInfinite' from Mathlib, resulting in a more readable proof.}
\label{fig:ragnrag1}
\end{figure}

\begin{figure}[tbp]
    \begin{subfigure}{0.45\textwidth}
    \vtop{
        \begin{lstlisting}
example {G H : Type*} {p : ℕ} [Group G] [Group H] [Fact p.Prime] (gp : IsPGroup p G) (f : G →* H) (sf : Function.Surjective f) : IsPGroup p H := by
  exact IsPGroup.of_surjective gp f sf
        \end{lstlisting}        
        \caption{Proof with LeanSearch-PS.}
        \label{fig:with_leansearch}
    }
    \end{subfigure}
    \hspace{0.02\textwidth}
    \begin{subfigure}{0.48\textwidth}
    \vtop{
        \begin{lstlisting}        
example {G H : Type*} {p : ℕ} [Group G] [Group H] [Fact p.Prime] (gp : IsPGroup p G) (f : G →* H) (sf : Function.Surjective f) : IsPGroup p H := by
  rw [IsPGroup] at gp ⊢
  intro g
  obtain ⟨m, hm⟩ :=sf g
  rcases gp m with ⟨km, hkm⟩
  use km
  rw [← hm, ← map_pow, hkm, map_one f]
        \end{lstlisting}        
        \caption{Proof without LeanSearch-PS.}
        \label{fig:without_leansearch}
    }
    \end{subfigure}
\caption{Compare the proofs with and without LeanSearch-PS. The prover in (a) uses the existing instance `IsPGroup.of\_surjective' from Mathlib, resulting in a more readable proof.}
\label{fig:ragnrag2}
\end{figure}

\end{document}